\documentclass[review]{elsarticle}

\usepackage{hyperref,multirow,amsmath,xcolor,ulem}

\journal{Journal of Pattern Recognition}

\begin{document}

\begin{frontmatter}

\title{A pooling based scene text proposal technique for scene text reading in the wild}


\author[label1,label6]{Dinh NguyenVan\corref{corAuthor}}
\address[label1]{Sorbonne University - University Pierre and Marie CURIE\\
				 4 place Jussieu 75252 Paris cedex 05, France\\
                 dinh.nguyen\_van@upmc.etu.fr\\
                 nizar.ouarti@ipal.cnrs.fr}

\author[label2]{Shijian Lu}
\address[label2]{Nanyang Technological University\\
                 Nanyang Avenue, Singapore 639798 \\
                 Shijian.Lu@ntu.edu.sg
                 }
                 
\author[label4]{Shangxuan Tian}
\address[label4]{Tencent Co. LTD, China\\
				Gaoxinnanyi Avenue,Southern District of Hi-tech Park,Shenzhen,China,518057\\
				sxtian@tencent.com
                }                 

\author[label1,label6]{Nizar Ouarti}

\author[label5,label6]{Mounir Mokhtari}
\address[label5]{Institut Mines-Telecom\\
                 37-39 Rue Dareau, 75014 Paris, France\\                 
                 Mounir.Mokhtari@mines-telecom.fr}
\address[label6]{Image \& Pervasive Access Lab (UMI 2955), CNRS, I2R\\
                 1 Fusionopolis Way, \#21-01 Connexis (South Tower), Singapore 138632}
%
%

\begin{abstract}
Automatic reading texts in scenes has attracted increasing interest in recent years as texts often carry rich semantic information that is useful for scene understanding. In this paper, we propose a novel scene text proposal technique aiming for accurate reading texts in scenes. Inspired by the pooling layer in the deep neural network architecture, a pooling based scene text proposal technique is developed. A novel score function is designed which exploits the histogram of oriented gradients and is capable of ranking the proposals according to their probabilities of being text. An end-to-end scene text reading system has also been developed by incorporating the proposed scene text proposal technique where false alarms elimination and words recognition are performed simultaneously. Extensive experiments over several public datasets show that the proposed technique can handle multi-orientation and multi-language scene texts and obtains outstanding proposal performance. The developed end-to-end systems also achieve very competitive scene text spotting and reading performance.
\end{abstract}


\end{frontmatter}

\section{Introduction}
Texts in scenes play an important role in communication between a machine and its surrounding environments. Automated machine understanding of texts in scenes has a vast range of applications such as navigation assistance for the visually impaired \cite{OrCam}, scene text translation for tourists \cite{GoogleTrans}, etc. It has been a grand challenge for years in the computer vision research community, and it received increasing interests in recent years as observed by a number of scene text reading competitions \cite{ICDAR2013,IncidentialText,SVT,ChineseICDAR}.
\par

Automated reading texts in scenes is a very challenging task due to the large intra-class variations and  the small inter-class variations with respect to many non-text objects in scenes. In particular, scene texts often have very different appearance because they may be printed in very different fonts and styles, captured from different distances and perspectives, and have very different cluttered background and lighting conditions. At the same time, texts often have high resemblance to many non-text objects in scenes, e.g. letter 'o'/'O' has similar appearance as many circular objects such as vehicle wheels, letter 'l' has similar appearance as many linear structures such as poles, etc. 
\par 

In this research, we design a novel scene text proposal technique and integrate it into an end-to-end scene text reading system. Inspired by the pooling layer in deep neural network, a pooling based scene text proposal technique is designed which is capable of grouping image edges into word and text line proposals efficiently. One unique feature of the pooling based proposal technique is that it does not involve heuristic thresholds/parameters such as text sizes, inter-character distances, etc. that are often used in many existing scene text detection techniques \cite{TextFlow,Shij,GVF,TDCNN7}. A novel score function is designed which exploits the histogram of oriented gradients and is capable of ranking proposals according to their probabilities of being text. Preliminary study on max-pooling based proposals has been presented in our prior work \cite{MPT_ICDAR2017}. This paper presents a more comprehensive study by investigating several key proposal parameters such as proposal quality, optimal proposal set-ups, etc. In addition, several new studies are performed from different aspects of edge label assignments, pooling methods and detection of arbitrarily oriented scene texts in different languages. Furthermore, a new score function is designed which is more efficient and robust in proposal ranking. We also integrate the proposed pooling based proposal technique into an end-to-end scene text reading system and study its effects on the scene text reading performance.
\par

The rest of this paper is organized as follows. Section 2 reviews recent works in scene text proposals, scene text detections, and scene text reading. Sections 3 and 4 describe the proposed scene text proposal technique and its application in scene text reading. Section 5 presents experimental results and several concluding remarks are drawn in Section 6.

\section{Related works}
Traditional scene text reading systems consist of a scene text detection step and a scene text recognition step. In recent years, scene text proposal has been investigated as an alternative to the scene text detection, largely due to its higher recall rate which is capable of locating more text regions as compared with the traditional text detection step.

\subsection{Scene text proposal}
The scene text proposal idea is mainly inspired by the success of object proposal in many object detection systems. It has advantage in locating more possible text regions to offer higher detection recall. It's often evaluated according to the recall rate as well as the number of needed proposals - typically the smaller the better at a similar recall level \cite{HowGoodPP}. False-positive scene text proposals are usually eliminated by either a text/nontext classifier \cite{STL,TDCNN3DT} or a scene text recognition model \cite{GomezE2E,TEB} in end-to-end scene text reading systems.
\par

Different scene text proposal approaches have been explored. One widely adopted approach combines generic object proposal techniques with text-specific features for scene text proposal generation. For example, EdgeBoxes \cite{EB} is combined with two text-specific features for scene text proposal generation \cite{TEB}. In another work \cite{JarE2E}, EdgeBoxes is combined with the Aggregate Channel Feature (ACF) and AdaBoost classifiers to search for text regions. In \cite{GomezE2E}, Selective Search \cite{SS} is combined with Maximally Stable Extremal Regions (MSER) to extract texture features for dendrogram grouping. A text-specific symmetry feature is explored in \cite{STL} to search for text line proposals directly, where false text line proposals are removed by training a CNN classifier. Deep features have also been used for scene text proposal due to its superior performance in recent years. For example, inception layers are built on top of the last convolution layer of the VGG16 for generating text proposal candidates in \cite{TDCNN3DT}. The Region Proposal Network (RPN) and Faster R-CNN structure are adopted for scene text proposal generation in \cite{TDCNN7,TextRCNN}.
\par

Most existing scene text proposal techniques have various limitations. For example, the EdgeBoxes based technique \cite{JarE2E} is efficient but often generate a large number of false-positive proposals. The hand-crafted text-specific features rely heavily on object boundaries which are sensitive to image noise and degradation \cite{Shij}. Techniques using heuristic rules and parameters \cite{TEB} do not adapt well across datasets. The deep learning based technique \cite{TDCNN3DT} produces a small number of proposals but the recall rate becomes unstable when the Intersection over Union (IoU) threshold increases. As a comparison, our proposed proposal technique does not leverage heuristic parameters and obtains a high recall rate with a small number of false-positive proposals.
 
\subsection{Scene text detections}
A large number of scene text detection techniques have been reported in the literature. Sliding window has been widely used to search for texts in scene images \cite{TextFlow,Coarse2FineConv,TaoWang}. However, it usually has a low efficiency because it adopts an exhaustive search process by using multiple windows of different sizes and aspect ratios. Region based techniques have been proposed to overcome the low efficiency constraint. For example, the Maximal Stable External Regions (MSRE) has been widely used \cite{Re1,Re21,Re5,Re13} for scene text detection. In addition, various hand-craft text-specific features have also been extensively investigated such as Stroke Width Transform (SWT) \cite{Re2}, Stroke Feature Transform (SFT) \cite{Re8}, text edge specific features \cite{Shij}, Stroke End Keypoints (SEK), Stroke Bend Keypoints (SBK) \cite{Re20}, and deep features based regions \cite{TDCNN1,TDCNN6,TDCNN8}. Different post-processing schemes have also been designed to remove false positives, e.g heuristic rules based classifier \cite{GVF,Re13,Re15,Re18}, graph processing \cite{TextFlow,Coarse2FineConv}, support vector regression \cite{Shij}, convolutional K-mean\cite{Coarse2FineConv}, distance metric learning \cite{Re1}, AdaBoost \cite{Re5,Re14}, random forest \cite{Re2,Re8}, convolution neural network \cite{TaoWang,Re21}, etc.
\par

With the advance of convolutional neural network (CNN), different CNN models have been exploited for the scene text detection tasks. For example, the DeepText makes use of convolutional layers for deep features extraction and inception layers for bounding boxes predictions \cite{TDCNN3DT} . The TextBoxes \cite{TextBoxes} adopts the Single Shot Multiboxex Detector (SSD) \cite{SSD} to deal with multi-scale texts in scenes. Quadrilateral anchor boxes have also been proposed for detecting tighter scene text boxes \cite{DeepMatchPriorNet}. In addition, direct regression solution has also been proposed \cite{DeepDirectRegress} to remove the hand-crafted anchor boxes. Different CNN based detection and learning schemes have also been explored. For example, some work adopts a bottom-up approach that first detection characters and then group them to words or text lines \cite{TDCNN7,TDCNN4,WordSup}. Some system instead defines a text boundary class for pixel-level scene text detection \cite{WordFence, SelfOrg}. In addition, weakly supervised and semi-supervised learning approach \cite{WeakNet} has also been studied to address the image annotation constraint \cite{WeText}.

\subsection{End-to-end scene text reading}
End-to-end scene text reading integrates detection and recognition into the same system to read texts in scenes. One popular system is a Google-Translation \cite{E2E5} which performs end-to-end scene text reading by integrating a list of techniques including three scene text detection methods, three scene text segmentation and grouping methods, two scene text recognition models, and language models for post-processing. In \cite{E2E4}, sliding window is combined with Histogram of Oriented Gradient feature extraction and Random Ferns Classifier to compute text saliency maps where words are extracted using External Regions (ER) and further re-scored using Support Vector Machine (SVM). In \cite{E2E3}, Adaboost and SVM text classifiers are applied on the extracted text regions using ER to localize scene texts which are further recognized under an Optical Character Recognition (OCR) framework. Similar approach was also adopted in \cite{E2E7}, where Maximal Stable External Regions (MSER) instead of ER is implemented for scene text region localization. In \cite{E2E9}, Stroke Width Transform \cite{Re2} is adopted for scene text region detection and Random Forest is used for character recognition and words are further recognized by component linking, word partition, and dictionary based correction. In \cite{GomezE2E,JarE2E}, potential text regions are first localized using EdgeBox (EB) \cite{EB} or adapted simple selective search for scene text \cite{GomezE2E} and scene texts are further recognized using Jarderberg's scene text recognition model \cite{JarData}.
\par

Quite a number of CNN based end-to-end scene text reading systems have been reported in recent years.  In \cite{TaoWang,E2E11}, a CNN based character recognition model is developed where word information is extracted from text saliency map using sliding windows. The same framework has been implemented in \cite{JarTReg}, where a more robust end-to-end scene text reading system is developed by training 
a model handling three functions including text and non-text classification, case-insensitive characters recognition, and case-sensitive characters recognition. In \cite{TextBoxes}, an advanced end-to-end scene text reading system is designed where the Single Shot Multiboxes Detector (SSD) is employed for scene text detection and a transcription model proposed in \cite{E2ETrainable}  is adopted for recognition. End-to-end trainable scene text reading system has also been proposed which can concurrently produce texts location and text transcription \cite{DTSpotter}  
\par 

Our developed end-to-end scene text reading system adopts a similar framework as presented in \cite{GomezE2E,JarE2E} that exploits proposals and existing scene text recognition models. One unique feature is that it uses only around one-fifth of the number of proposals that prior proposal based end-to-end systems use thanks to our proposed pooling based proposal technique and gradient histogram based proposal ranking.
	
\section{Pooling based scene text proposal}
\begin{figure}[t]
\begin{center}
\includegraphics[width=4.5in]{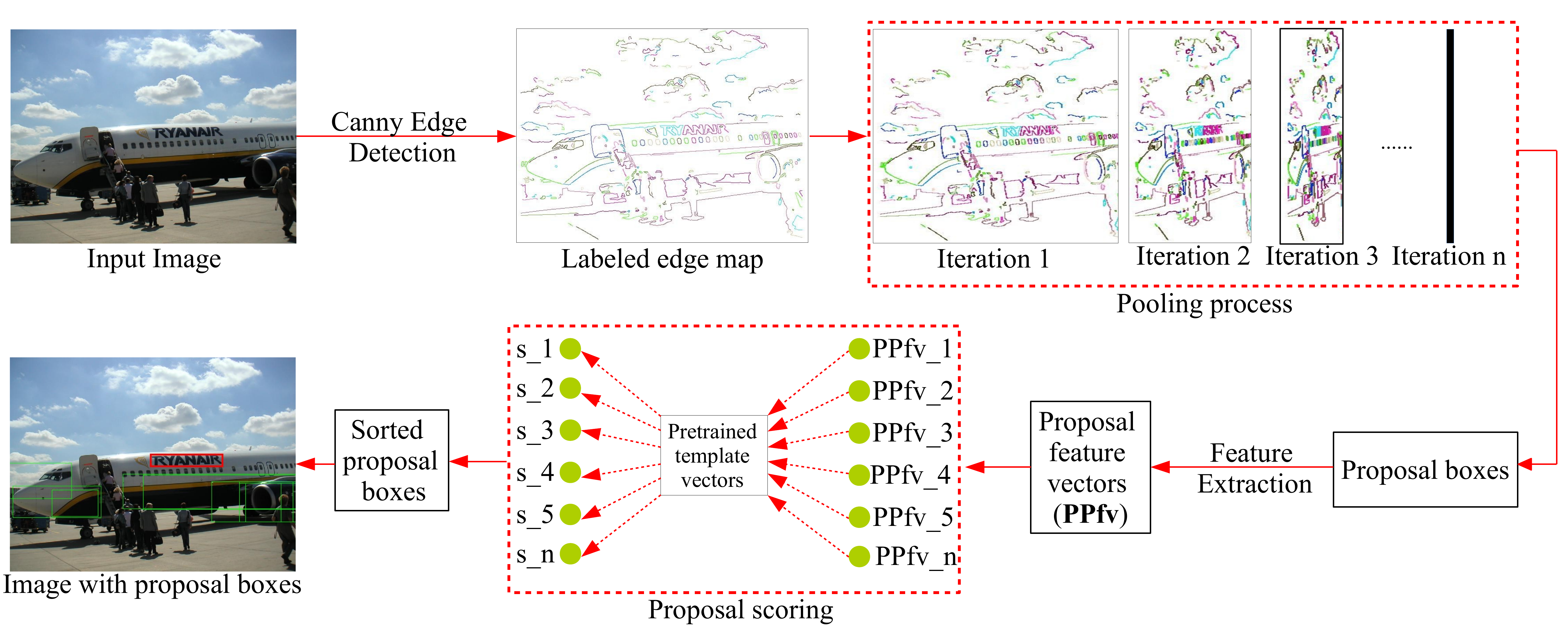}
\end{center}
\caption{The framework of the proposed scene text proposal technique including pooling based scene text proposal generation and proposal ranking based on low-level gradient features. In the output image, proposals are presented for the illustration purpose, where the proposal box in red colour detects the scene text correctly.}
\centering
\label{fig_001}
\end{figure}

The proposed scene text proposal technique follows the general object proposal framework which consists of two major steps including proposal generation and proposal ranking. For the proposal generation, we design a pooling based technique that iteratively groups image edges into possible words or text lines. Here each edge component could be a part of a single character, several neighbouring characters touching each other, or other non-text objects. Each set of grouped image edges thus forms a proposal which can be represented by a bounding box that covers all grouped edges. For proposal ranking, a scoring function is designed which is capable of ranking the determined proposals according to their probability of being text. The ranking strategy employs the histogram of oriented gradient which first learns a number of text and non-text templates and then ranks proposals according to their distances to the learned templates. Fig. \ref{fig_001} illustrates the framework of our proposed scene text proposal technique.

\subsection{Proposal generation}
\label{ppgens}
A novel pooling based technique is designed for the scene text proposal generation. The idea is inspired by the pooling layer in the convolution neural network (CNN) which is employed to eliminate insignificant features while shrinking a feature map. Given an image, an edge map is first determined by using the Canny edge detector \cite{Canny}, where each binary edge can be labelled through connected components (CC) analysis. Each binary edge can then be labelled by an unique number indicating the order when it is searched. For example, the first searched binary edge is assigned an unique label number 1, the second with an unique number 2, etc. An initial edge feature map can thus be determined by assigning all pixels of a binary edge with the same number as the component label and all non-edge pixels with a number of zero. 
\par

The image edge feature map is then processed iteratively through pooling using a pooling window. Take max-pooling as an example. During each pooling iteration, only the pixel with the largest label number within the pooling window is kept for generating the new edge feature map for the next iteration, and all other pixels with a smaller label number are discarded. The binary edges are therefore shifting to each others iteratively where those closer to each other are grouped first and those farther away are merged later. The iterative edge merging process terminates when there is no zero pixels existing in the edge feature map, meaning that there are no more gaps between the labelled binary edges as illustrated in the `Pooling process' in Fig. \ref{fig_001}. Multiple proposals are accordingly generated with different groups of edges throughout the pooling process.
\par

\begin{figure}[t]
\begin{center}
\includegraphics[width=4.5in]{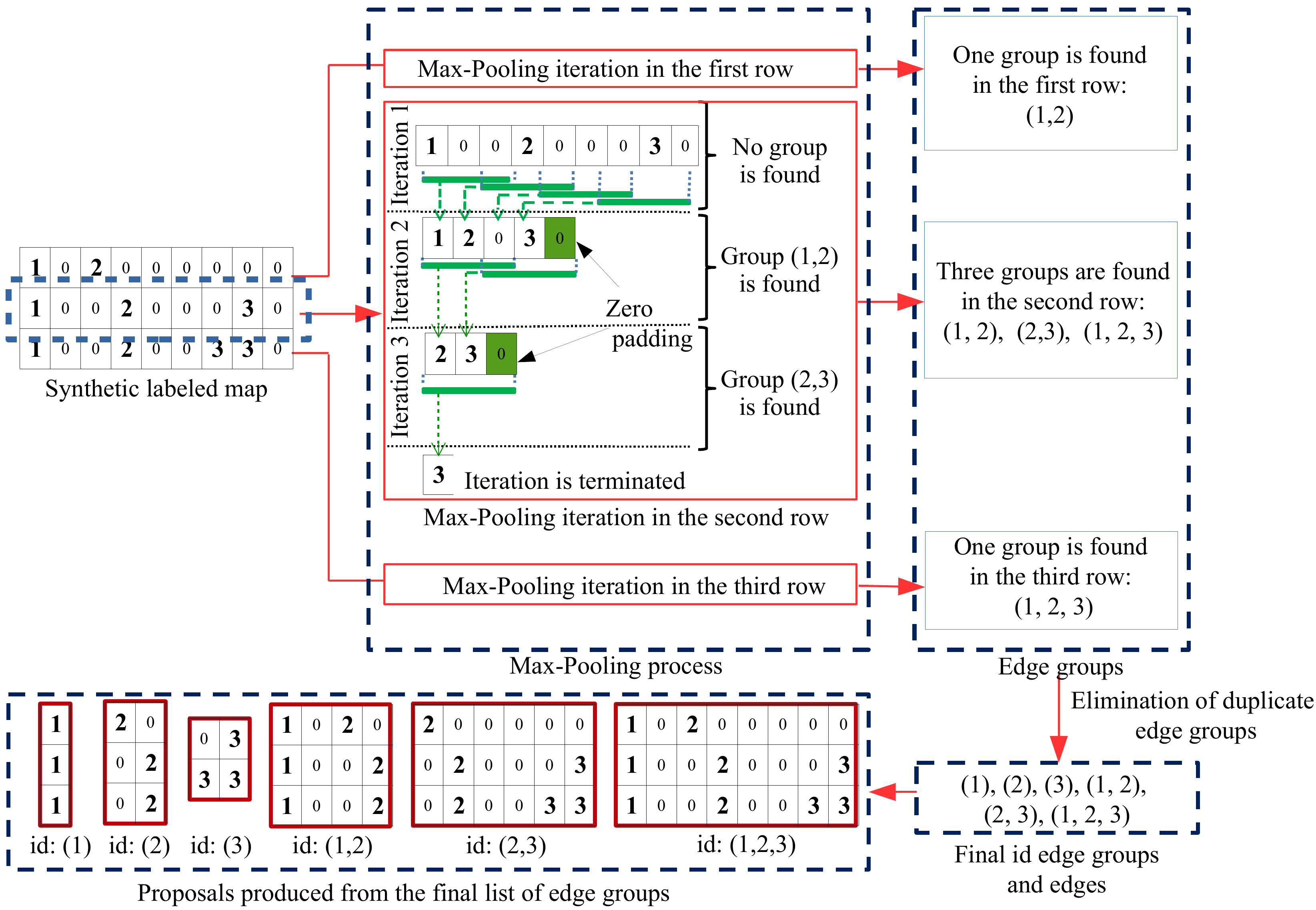}
\end{center}
\caption{Illustration of the max-pooling based scene text proposal generation with a 1-by-3 max pooling window and a horizontal stride of 2 on the synthetic labelled map - more detail on the second row are provided. Zero-padding is performed when the number of column is even. Duplicate proposals are removed, and six proposals are generated including three connected binary edges themselves and three found edge groups. 
}
\centering
\label{fig_002_1}
\end{figure}

\begin{figure}[t]
\begin{center}
\includegraphics[width=4.5in]{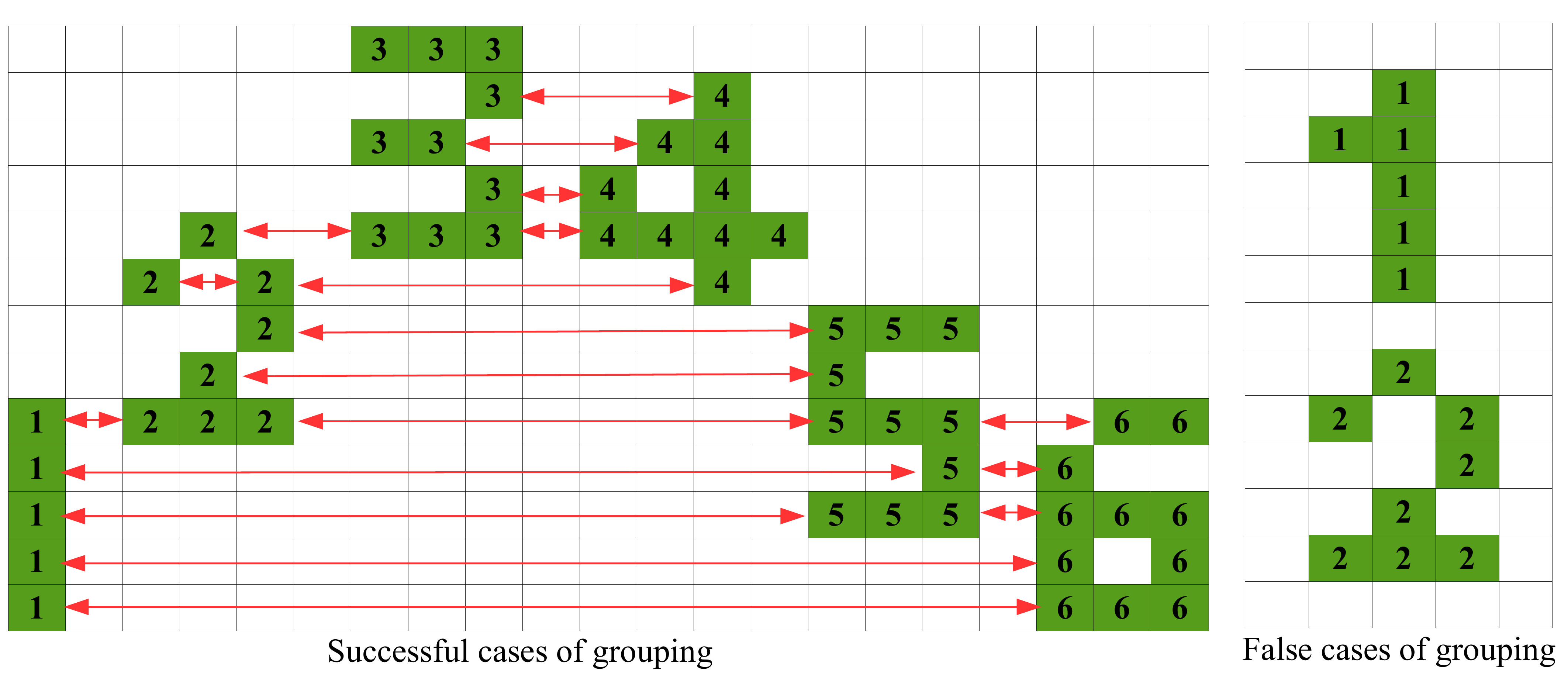}
\end{center}
\caption{
Two synthetic graphs that explain how horizontal pooling windows and horizontal stride can group non-horizontal text lines: The red-color two-directional arrows illustrate the link-up as determined during the pooling based grouping process. It shows that the proposed technique is able to handle non-horizontal text lines as far as the constituting letters/digits having overlapping in the vertical direction. The graph on the right illustrates when two neighbouring will not be grouped.
}
\centering
\label{fig_002}
\end{figure}
Fig. \ref{fig_002_1} illustrates the max-pooling based proposal generation process by using a synthetic edge map that contains 3 binary edges as labelled by 1, 2, and 3. Taking the second row as an example, the first two edges are grouped to form a proposal after the second pooling iteration and the second and third edges are grouped to form another proposal after the third pooling iteration. Since the first and the third edges are both grouped with the second edge, all three edges are also grouped to form a new proposal. For the first and the third row, a single group of the first and second edges and a single group of three single edges can be derived under the similar idea, respectively. By removing duplicated proposals, six proposals are finally determined including the three single edges, the grouped first and second edges, the grouped second and third edges, and the grouped three edges. It should be noted that zero-padding is implemented at the right side when the studied row has an even number of pixels left.
\par 

Though a horizontal pooling window of size 1-by-3 and a horizontal stride of 2 are used, the proposed pooling based proposal technique is able to handle non-horizontal words or text lines as far as the constituting letters/digits having certain overlap in the vertical direction. This is illustrated in the two synthetic graphs in Fig. \ref{fig_002}. As the first graph in Fig. \ref{fig_002} shows, the curved chain of digits 1-6 will be grouped together due to their overlap in the vertical direction. Note digits/letters could be grouped via other neighbouring digits/letters when they have no overlap in the vertical direction. For example, the digits 4 and 5 can be grouped via the digits 2 and 3 even though they have no vertical overlap. Digits/letters will not be grouped when they have no overlap in the vertical direction and also have no neighbouring digits/letters to leverage as illustrated in the second graph in Fig. \ref{fig_002}.

\subsection{Proposal ranking}
Histogram of oriented gradient has been used successfully for the scene text detection and recognition tasks \cite{Wang,Recog4}. The success shows that scene texts actually have certain unique HoG features that can differentiate them from other non-text objects. We therefore adapt HoG for proposal ranking, aiming to exploit the unique text-specific HoG features to rank text proposals to the front of the whole proposal list. Different from the traditional HoG, we extract HoG features from the Canny edge pixels only which we will refer it by Histogram of Oriented Gradient on edges (HoGe) in the ensuing discussion. 
\par

In our proposal ranking strategy, a number of text and non-text HoGe templates are first learned from a set of training images to be discussed in \ref{ranking optimization}. Scene text proposals are then scored and ranked according to the distances between their HoGe and the learned text and non-text HoGe templates. The scoring function is defined as follows:

\begin{equation}
	s=\frac{d_{nt}}{d_{nt}+d_t}
\end{equation}
where $d_t$ and $d_{nt}$ refer to the distances between the feature vector ($F$) of a detected proposal and the pre-determined text and non-text templates as follows.

\begin{equation}
\begin{split}
d_t = \sum_{i=1}^{n}\lVert (F,T_i) \rVert \\
d_{nt} = \sum_{i=1}^{n}\lVert (F,NT_i) \rVert 
\end{split}
\end{equation}

where $n$ denotes the number of text ($T$) and non-text ($NT$) templates, and $\lVert \cdot \rVert$ gives the Euclidean distance between $F$ and a text/non-text feature template. The score function in Eq. 1 is designed based on the observation that the feature vector of a text proposal is usually closer to text templates as compared with non-text templates. The feature vector of a text proposal will thus produce a small $d_t$ and a large $d_{nt}$ which further lead to a high text probability score.

\subsection{Discussion}
\label{optimization}
\subsubsection{Pooling and edge labelling}
\label{PoolingLabeling}
As described in Section \ref{ppgens}, we assign edge labels according to the searching order (from left to right column by column and from top to bottom in each column) and adopt the max-pooling to group text edges within the same line. On the other hand, the proposed technique can work with different edge label assignment and pooling methods. Two new tests are performed for verification. The first test studies two more edge labelling methods that assign edge labels by using the maximum and mean gradient of pixels within an CC, respectively (named by \textit{maxE} and \textit{meanE} in Table \ref{tablePoolingLabeling}). Take the use of \textit{meanE} as an example. It first calculates the mean gradient of each CC and then labels all edge pixels by using the calculated mean gradient directly. The second test studies the min-pooling method that keeps the smallest instead of the largest edge labels (as in max-pooling) falling within the same pooling window.
\par

\begin{table}[!t]
\small
\caption{Recall rates of four variants of the proposed technique on test images of the ICDAR2003 and ICDAR2013 dataset under IoU threshold of 0.8 and four sets of pooling window sizes and stride values in the format of \textit{window height - window width - vertical stride - horizontal stride}.}
\begin{center}
\begin{tabular}{c c c c c}
\hline
\hline
    & 1-3-1-2 & 2-3-2-2 & 3-3-2-2 & 3-3-3-3 \\
\hline
\hline
\textit{maxE$\_$maxP} &  78.84 & 77.07  & 71.16 & 56.63 \\
\textit{meanE$\_$maxP}&  79.02 & 76.79  & 71.93 & 56.58 \\
\textit{searL$\_$maxP}&  79.52 & 76.11  & 71.12 & 55.5  \\
\textit{searL$\_$minP}&  78.93 & 76.52  & 71.39 & 56.18 \\
\hline
\hline
\end{tabular}
\end{center}
\label{tablePoolingLabeling}
\end{table}

Table \ref{tablePoolingLabeling} shows the test results on the test images of the ICDAR2003 and ICDAR2013 datasets, where \textit{searL} denotes a labelling method that assigns edge labels according to the edge searching order, \textit{maxP} and \textit{minP} denote max-pooling and min-pooling, respectively. So \textit{maxE$\_$maxP} means that image edges are labelled by using the maximum gradient and pooling is performed by max-pooling. The very close proposal recalls under different window sizes and strides in Table \ref{tablePoolingLabeling} verify that our proposed technique is tolerant to both edge labelling methods and edge label pooling methods. 
   
\subsubsection{Proposal generation}
\label{proposal generation optimization}

\begin{figure}[t]
\begin{center}
\includegraphics[width=4.5in]{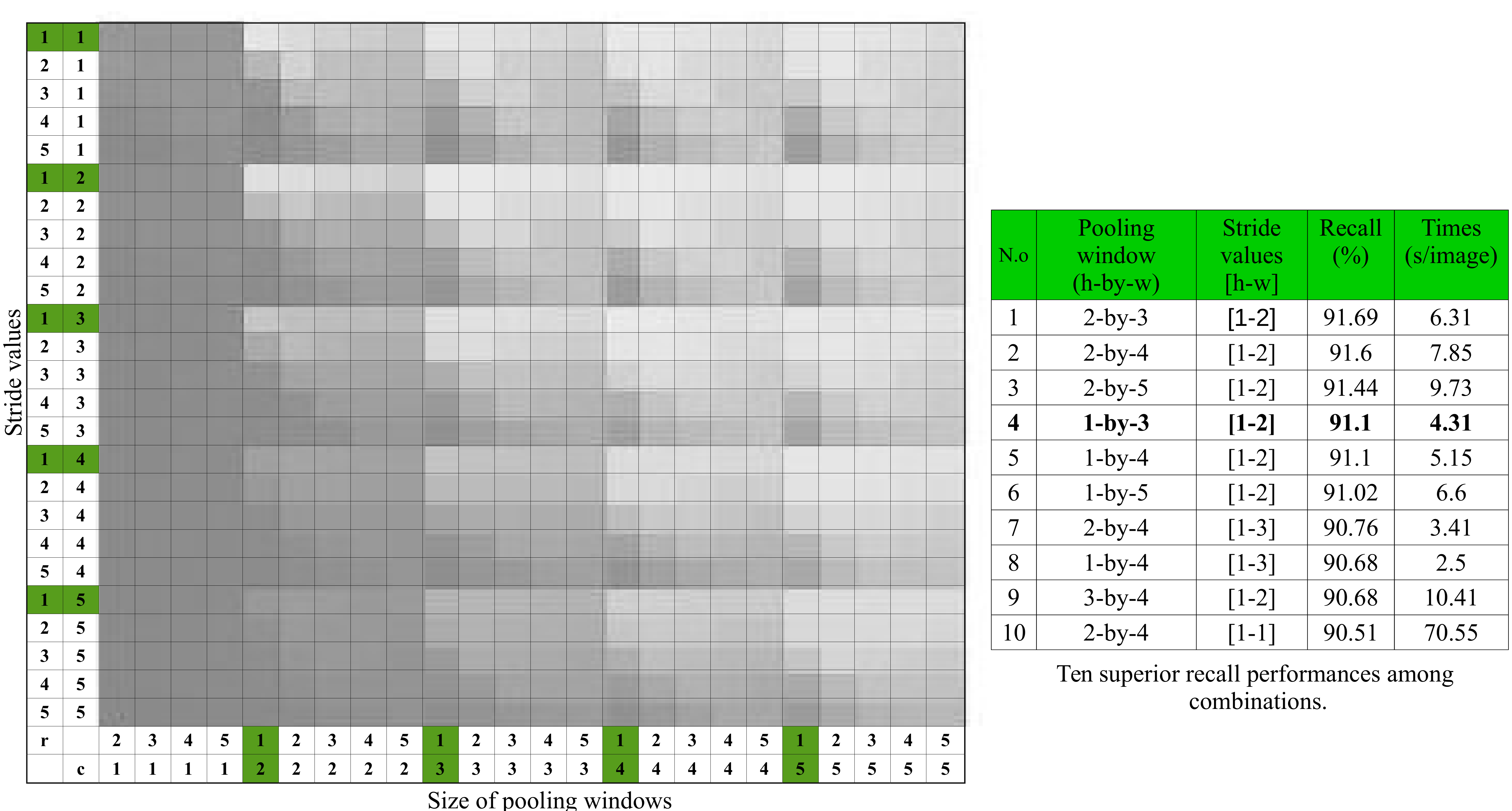}
\end{center}
\caption{The heat-map presents performances of the proposed proposal technique under different combinations of the pooling window size and the stride values, as evaluated on the ICDAR2003 and ICDAR2013 training datasets. This study shows that the optimal combination is a horizontal pooling window of 1-by-3 and a horizontal stride of 2 for a trade-off of accuracy and processing time}
\centering
\label{fig_008}
\end{figure}

As the optimization of proposal generation targets the best proposal recall, we relax the number of proposals and include all generated proposals while studying the size of the pooling window and strides. We adopt the grid search to study the two key sets of parameters, including a pooling window size (width and height) and stride values (a horizontal stride and a vertical stride). In particular, we vary the size of the pooling window and strides from 1 to 5 which produces 600 (24*25) parameter settings. Note the pooling window size 1-by-1 is not included as it does not perform any grouping operations. 
\par

Fig. \ref{fig_008} shows proposal recalls under the 600 parameters setting which are presented by using a heat-map, where each recall is an average of three recalls when three IoU thresholds 0.5, 0.7, and 0.8 are applied. As shown in Fig. \ref{fig_008}, the two numbers at the bottom of each column refer to the row number (the number at the top) and the column number (the number at the bottom) of the pooling window, respectively. The two numbers at the left of each row refer to strides in the vertical direction (the number on the left) and horizontal direction (the number on the right), respectively. We further sort the recalls under the 600 settings and the table on the right shows several best-performing settings. In our implemented system, we take a compromise between recall rate and processing time and select the combination of a 1-by-3 pooling window and strides 1-by-2 in vertical and horizontal directions.
\par

\begin{figure}[t]
\begin{center}
\includegraphics[width=3.5in]{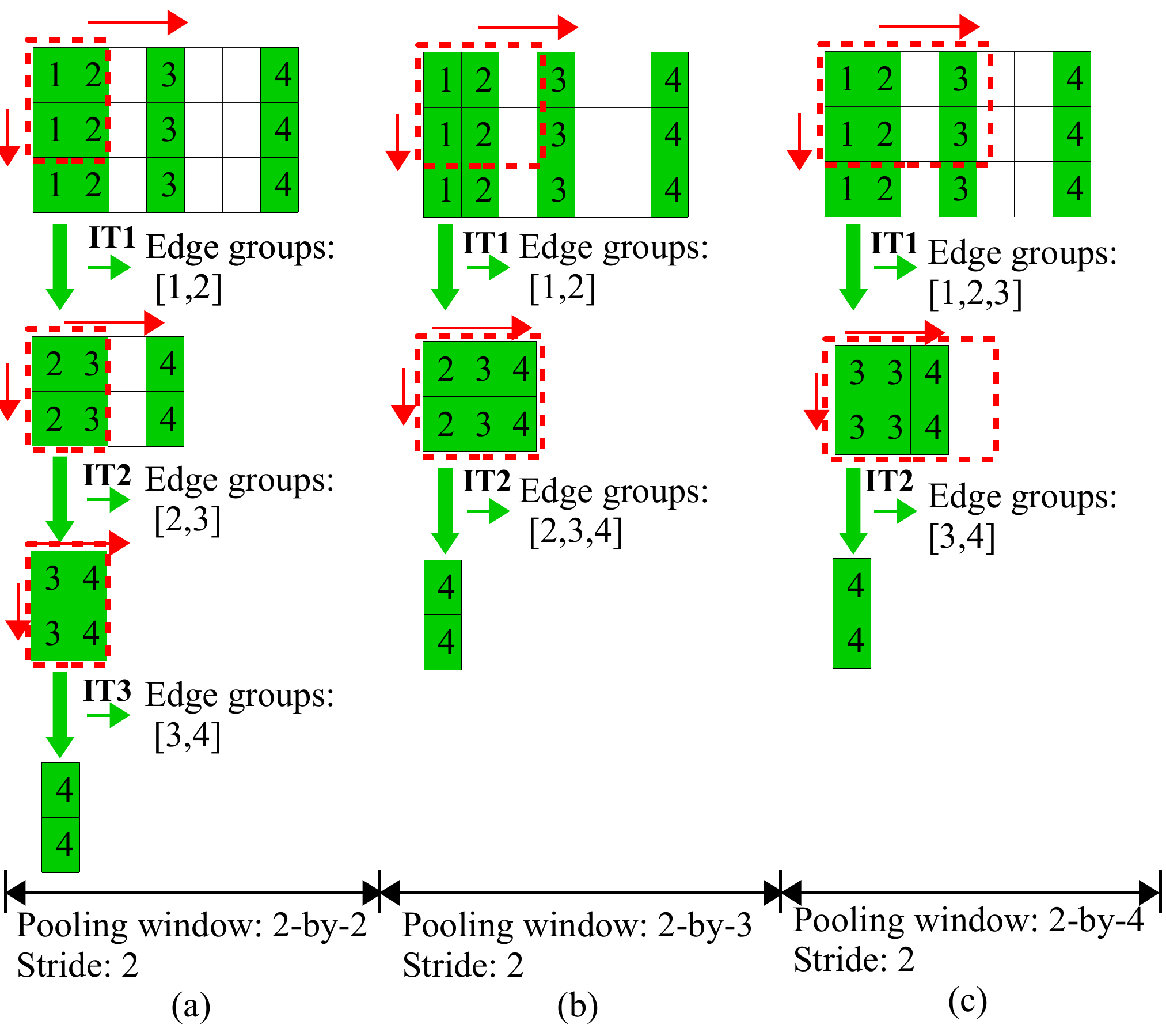}
\end{center}
\caption{The effect of pooling window size on the quality of proposals: Generally, a smaller pooling window will generate the larger number of proposals and have a better chance of capturing the right objects. The number of proposals decreases when pooling window size increases from graph (a) to graph (d) where \textbf{IT} denotes iteration.}
\centering
\label{fig_009}
\end{figure} 

Several factors need to be taken into consideration while setting the pooling and strides. The first is the absolute size of the pooling window which defines the minimum distance between neighbouring edges that the pooling based proposal technique could capture. For example, a large pooling window of size 2x4 will not be able to captures distances of 1, 2 and 3 pixels between neighbouring edges whereas a pooling window of size 2x2 is capable of capturing distance as small as 1 pixel only as illustrated in Fig. \ref{fig_009}. The second is specific setting of rows and columns of the pooling window and strides in horizontal and vertical directions. In particular, the increase of coverage/jump in the vertical direction often deteriorate the proposal performance as illustrated in the heat-map in Fig. \ref{fig_008}. One reason could be due to the fact that most text in scenes are positioned in a horizontal direction. In addition, a pooling window with a big span in vertical direction often groups texts with neighbouring non-text objects lying above or below texts. The third is overlap between two consecutive pooling windows which happens when the stride in the horizontal direction is smaller than the width of the pooling window. A smaller stride often produces better recall rate, meaning that overlaps between two consecutive pooling windows helps to produce better proposals. In fact, the proposal performance drops a lot when there are absolutely no overlaps between two consecutive pooling windows as illustrated in the heat-map in Fig. \ref{fig_008}.  

\subsubsection{Proposal ranking}
\label{ranking optimization}
 
\begin{table}[!t]
\small
\caption{Recall rates of the proposed scene text proposal technique (on the training images of the ICDAR2013 and the ICDAR2003 datasets) under different combinations of the number of templates per class (nC) and the template dimension (Dims).}
\begin{center}
\begin{tabular}{c c c c c c c}
\hline
\hline
No. & nC & Dims & Recall\_Aver & Recall\_IoU0.5 & Recall\_IoU0.7 & Recall\_IoU0.8 \\
\hline
\hline
1   &  25 & 120 & 88.75 & 94.96 & 89.67 & 81.61 \\[-0.05cm]
2   &  35 & 120 & 88.75 & 94.71 & 89.67 & 81.86 \\[-0.05cm]
3   &  65 & 140 & 88.75 & 94.71 & 89.67 & 81.86 \\[-0.05cm]
4   &  65 & 170 & 88.75 & 95.21 & 89.42 & 81.61 \\[-0.05cm]
5   &  25 & 180 & 88.75 & 95.21 & 89.42 & 81.61 \\[-0.05cm]
6   &  50 & 130 & 88.66 & 94.71 & 89.67 & 81.61 \\[-0.05cm]
7   &  50 & 140 & 88.66 & 94.71 & 89.67 & 81.61 \\[-0.05cm]
8   &  35 & 180 & 88.66 & 95.21 & 89.42 & 81.36 \\[-0.05cm]
9   &  45 & 120 & 88.58 & 94.96 & 89.42 & 81.36 \\[-0.05cm]
10  &  20 & 130 & 88.58 & 94.96 & 89.42 & 81.36 \\[-0.05cm]
\hline
\hline
\end{tabular}
\end{center}
\label{table5}
\end{table} 

We adopt a grid search strategy to investigate the optimal HoGe dimension and the number of text and non-text templates. The dimension of the HoGe feature vector refers to the number of histogram bins within the HoGe which we change from 10 to 180 with a step of 10. The number of text and no-text templates is varied from 5 to 100 with a step of 5. Hence, the full combination of the two sets of parameters thus gives 360 (18x20) settings. Different from the proposal generation optimization, we limit the maximum proposal number at 2000 (a reasonable number by compromising recall and the ensuing computational cost \cite{ObjDT}) for the evaluation of proposal recalls. Under each parameter setting, an average recall is computed for all images within the validation set when three IoU thresholds of 0.5, 0.7 and 0.8 are used. Additionally, 80\% training images of the ICDAR2003 and the ICDAR2013 datasets are used for training and the rest 20\% are used for validation in our study.
\par

Table \ref{table5} shows the first ten best-performing settings of the two parameters which are sorted according to the average recall under the three IoU thresholds. As Table \ref{table5} shows, the recalls are quite close to each other around the best parameter settings. In our implemented system, we select the 25 text/non-text templates and template dimension of 120, i.e., the setting (25, 120), as a compromise of detection recall and detection efficiency.

\section{Automatic scene text reading}
We also develop an end-to-end scene text reading system by integrating the proposed pooling based proposal technique and a state-of-the-art scene text recognition model \cite{JarData} which is trained on generic 90k words list and recognizes words directly. Given an image, a number of scene text proposals are first determined by using the proposed pooling based technique. Each detected proposal is then fed to the word recognition model \cite{JarData} to derive a word recognition score, and it will be discarded if the recognition score is too low or the recognized word is not in the lexicon list. After that, non-maximum-suppression (nms) is applied to keep the proposal with the maximum score and remove those with lower scores. Additionally, a word based nms is also implemented to remove duplicate proposals of the same word. In particular, only a proposal that has the maximum recognition score is kept as the reading output when more than one proposals overlap with each other and produce the same recognized word. More details will be discussed in Section \ref{E2EEval}.

\section{Experiments and results}
\subsection{Experiment setup and evaluation metrics}
Given the very similar performance under different label assignment and pooling methods as described in Section \ref{PoolingLabeling}, we label image edges by their searching order and use the max-pooling in the ensuing evaluations and benchmarking with the state-of-the-arts. In addition, the size of the pooling window is fixed at 1-by-3 and the strides are set at 2 and 1 pixels in the horizontal and vertical directions as described in Section \ref{proposal generation optimization}. Further, 25 feature templates are used for both text and non-text classes and the dimension of the HoGe is fixed at 120 bins as discussed in Section \ref{ranking optimization}.
\par

Three datasets are used in evaluations and comparisons including the focused scene text dataset used in the Robust Reading Competition 2015 (ICDAR2015) \cite{ICDAR2013}, the Street View Text (SVT) \cite{SVT} and the MSRA-TD500 \cite{Re3}. The ICDAR2015 contains 229 training images and 233 testing images, and the SVT contains 101 training images and 249 testing images. The scene text images in both datasets suffer from a wide range of image degradation but most texts are horizontal and printed in English. The MSRA-TD500 contains 500 images including 300 training images and 200 test images, where scene texts are in arbitrary orientations and a mixture of English and Chinese. It is used to show that the proposed technique can work with scene texts in different orientations and languages.
\par

The proposal quality is evaluated by the recall rate, the number of proposal selected and the computation time. The criterion is that a better proposal technique is capable of achieving a higher recall rate with a smaller number of proposals and a lower computation cost. While benchmarking different proposal techniques, the recall rate can be compared by fixing the number of proposals, says 2000 as a widely adopted number \cite{ObjDT}. In addition, the recall rate is also affected by the IoU threshold where a larger IoU usually leads to a lower recall rate. For the scene text reading system, two evaluation criteria are adopted as used in the robust reading competitions \cite{Wang}, namely, the 
end-to-end based and the spotting based. The end-to-end based evaluation focuses on alphanumeric words, while the spotting based evaluation targets words consisting of letters only. In particular, a correct word should have at least three characters (otherwise ignored), and only proposals that have over 50\% overlap with corresponding ground truth boxes and contain correctly recognized words are counted as true positives.

\subsection{Comparisons with state-of-the-arts}
The proposed technique (MPT) is compared to several state-of-the-art scene text proposal techniques including Simple Selective Search for Text Proposal (TP) \cite{GomezE2E}, Symmetry Text Line (STL) \cite{STL}, and DeepText (DT) \cite{TDCNN3DT}. In addition, we also compare the MPT with several state-of-the-art generic object proposal methods including EdgeBox (EB) \cite{EB}, Geodesic (GOP) \cite{GOP}, Randomized Prime (RP) \cite{RP}, and Multiscale Combination Grouping (MCG) \cite{MCG}. All these techniques are implemented in Matlab except TP and STL which are implemented in C++. All evaluations are performed on a HP workstation with a Intel Xeon 3.5GHz x 12 CPU and 32GB Ram memory. 

\begin{figure}[!t]
\begin{center}
\includegraphics[width=4.7in]{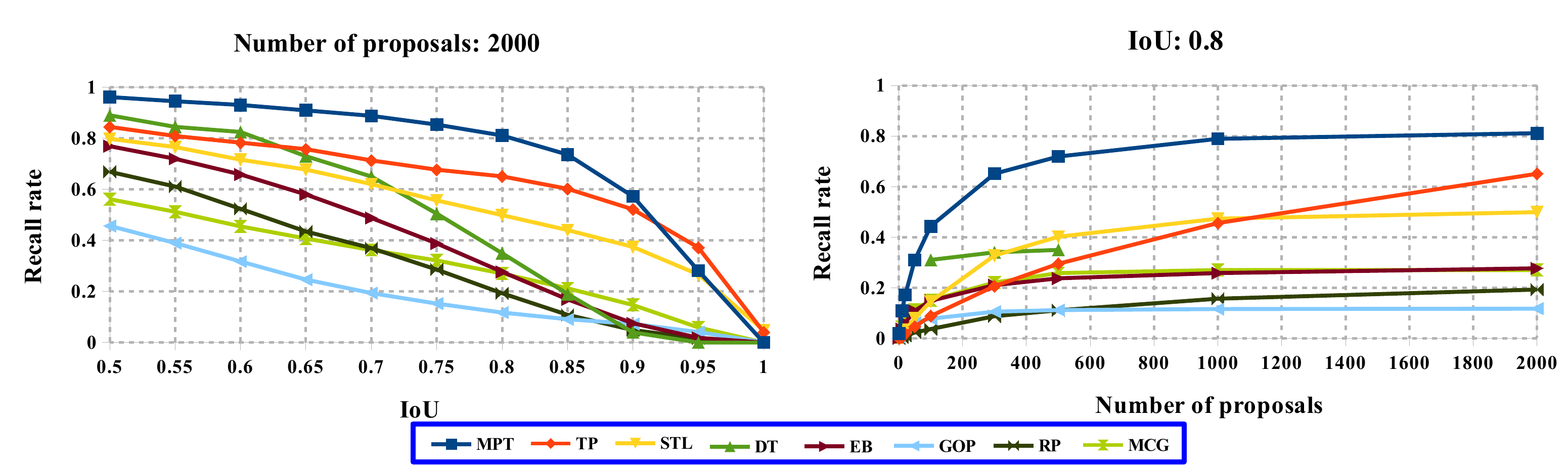}
\end{center}
\caption{Comparison of the proposed Max-pooling based scene text proposal technique (MPT) with state-of-the-art technique including text-specific proposal techniques: Simple Selective Search for Scene Text Proposal (TP) \cite{GomezE2E}, Symmetry Text Line (STL) \cite{STL}, and Deep Text (DT) \cite{TDCNN3DT} and generic object proposal techniques: EdgeBoxes (EB) \cite{EB}, Geodesic (GOP) \cite{GOP}, Randomized Prime (RP) \cite{RP}, and Multiscale Combination Grouping (MCG) \cite{MCG}. The evaluation is performed on the ICDAR2015 dataset by varying the IoU and the number of selected proposals.}
\centering
\label{fig_004}
\end{figure}
\par 

Fig. \ref{fig_004} shows experimental results on the ICDAR2015 dataset. The graph on the left shows proposal recalls when the IoU thresholds changes from 0.5 to 1 with a step of 0.05 and 2000 proposals selected from each image. The graph on the right shows recalls while the number of proposals varies from 1 to 2000 when the IoU threshold is fixed at 0.8. As the graph on the left shows, DT demonstrates competitive recalls when the IoU threshold lies between 0.5 and 0.6, but its recall drops dramatically when the IoU threshold increases. TP and STL are stabler than DT as they both use hand-craft text specific features, but their recalls are lower than the proposed MPT except when the IoU threshold is large than 0.9, which is seldom adopted in real systems. In the right graph, the proposed MPT outperforms most compared techniques when the number of proposals changes. In fact, it even outperforms DT which adopts a deep learning approach. Note that the recalls of DT are only evaluated in the range of 100-500 proposals because it set the maximum proposal number at 500.
\par 

\begin{table}[!t]
\small
\caption{Recall (\%) and processing time (in seconds) of the proposed technique (MPT) and state-of-the-art techniques under different IoU on the ICDAR2015 dataset. The Nppb denotes an average number of proposals each technique needs to achieve its presented recalls.}
\begin{center}
\begin{tabular}{c c c c c c}
\hline
\hline
Method & IoU: 0.5 & IoU: 0.7 & IoU: 0.8 & Nppb & times (s)  \\[-0.05cm]
\hline
\hline
MPT & \textbf{96.16}	 & \textbf{89.59} & \textbf{81.83} & 1465 &	  3.55\\[-0.05cm]

TP \cite{GomezE2E} & 84.47 & 71.32 & 65.11 & 1907 & 5.17\\[-0.05cm]

STL \cite{STL} & 79.78 & 62.04 & 49.91 & 1034& 361.3\\[-0.05cm]

DT \cite{TDCNN3DT} & 88.5 & 67 & 4 & \textbf{500} & $-$\\[-0.05cm]

EB \cite{EB} & 76.93 & 48.81 & 27.67 & 1968& \textbf{1.02}\\[-0.05cm]

GOP \cite{GOP} & 45.68 & 19.39 & 11.76 &1040 & 4.3\\[-0.05cm]

RP \cite{RP} & 66.91 & 36.95 & 19.3 & 1917 & 10.07\\[-0.05cm]

MCG \cite{MCG} & 56.16 & 36.21 & 27.02 & 550 & 28.92\\[-0.05cm]
\hline
\hline
\end{tabular}
\end{center}
\label{table1}
\end{table}

\begin{table}[!t]
\small
\caption{Recall (\%) and processing time (in seconds) of the proposed technique (MPT) and state-of-the-art techniques under different IoU on the SVT dataset. The Nppb denotes an average number of proposals each technique needs to achieve its presented recalls.}
\begin{center}
\begin{tabular}{c c c c c c}
\hline
\hline
Method & IoU: 0.5 & IoU: 0.7 & IoU: 0.8 & Nppb & times (s)  \\[-0.05cm]
\hline
\hline
MPT & \textbf{87.64} & 46.48 & 20.87 & 1780 &  3.19\\[-0.05cm]

TP \cite{GomezE2E} & 74.65 & 42.5 & 21.33 & 1972 & 5.94\\[-0.05cm]

STL \cite{STL} & 77.13 & 31.07 & 10.36 & 1358 & 433.82\\[-0.05cm]

EB \cite{EB} & 76.35 & \textbf{47.45} & \textbf{23.96} & 2000 & \textbf{1.28}\\[-0.05cm]

GOP \cite{GOP} & 52.09 & 18.24 & 6.8 & 1117 & 3.78\\[-0.05cm]

RP \cite{RP} & 62.6 & 27.05 & 12.21 & 2000 & 8.02\\[-0.05cm]

MCG \cite{MCG} & 54.71 & 24.27 & 8.66 & \textbf{557} & 14.97\\[-0.05cm]
\hline
\hline
\end{tabular}
\end{center}
\label{table2}
\end{table}

We also studied the number of needed proposals for good recalls and computational cost. Tables \ref{table1} and \ref{table2} show the experimental results on the test images of the dataset ICDAR2015 and SVT. It can be seen that the proposed MPT outperforms other proposal techniques in most cases for both datasets. TP is also competitive but it requires a larger number of proposals and also higher computational cost. EB is the most efficient and MCG requires a smaller number of proposals but both methods have low recalls under different IoU thresholds.
\par 

\begin{figure}[t]
\begin{center}
\includegraphics[width=4.8in]{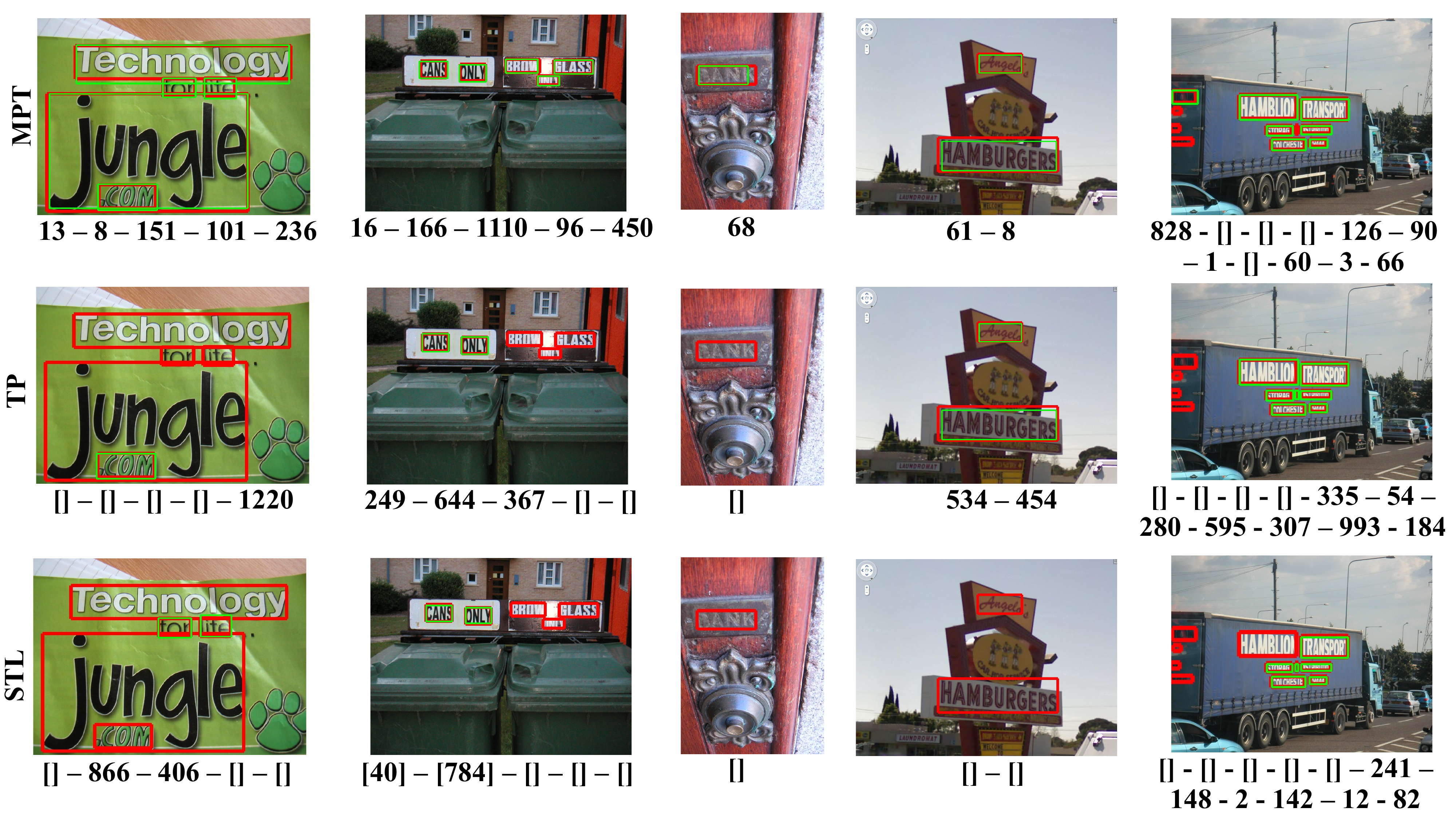}
\end{center}
\caption{
Performance of the proposed MPT and two state-of-the-art techniques STL and TP on several images in the ICDAR2015 and SVT datasets. The numbers under each image correspond to the position of proposals within the ranked proposal list (the smaller, the better). Note there are more than one good proposal in the proposal list for each ground truth box, and we only show the best proposal for each ground truth box for the illustration purpose.}
\centering
\label{fig_005}
\end{figure}

Fig. \ref{fig_005} illustrates the performance of the proposed MPT and compares it with two state-of-the-art techniques TP and STL (green boxes indicate proposals and red boxes indicate ground-truth). Several sample images are selected from the ICDAR2015 and SVT datasets which suffer from different types of degradations including text size variation (the first images from left), uneven illumination (the second image), ultra-low contrast (the third image), and perspective distortion (the fourth and fifth images). A series of numbers are shown under each image which correspond to the position of each proposal within the ranked proposal list (the smaller, the better). As Fig. \ref{fig_005} shows, the proposed MPT can deal with different types of image degradation and demonstrates superior proposal performance as compared with TP and STL. It should be noted that Fig. \ref{fig_005} only shows good proposals that have over 80\% overlap with ground-truth boxes. In addition, each text ground truth has more than one good proposal and Fig. \ref{fig_005} only shows the proposal which is ranked at the front-most with the smallest index number within the ranked proposal list.
\par

\begin{figure}[t]
\begin{center}
\includegraphics[width=4.8in]{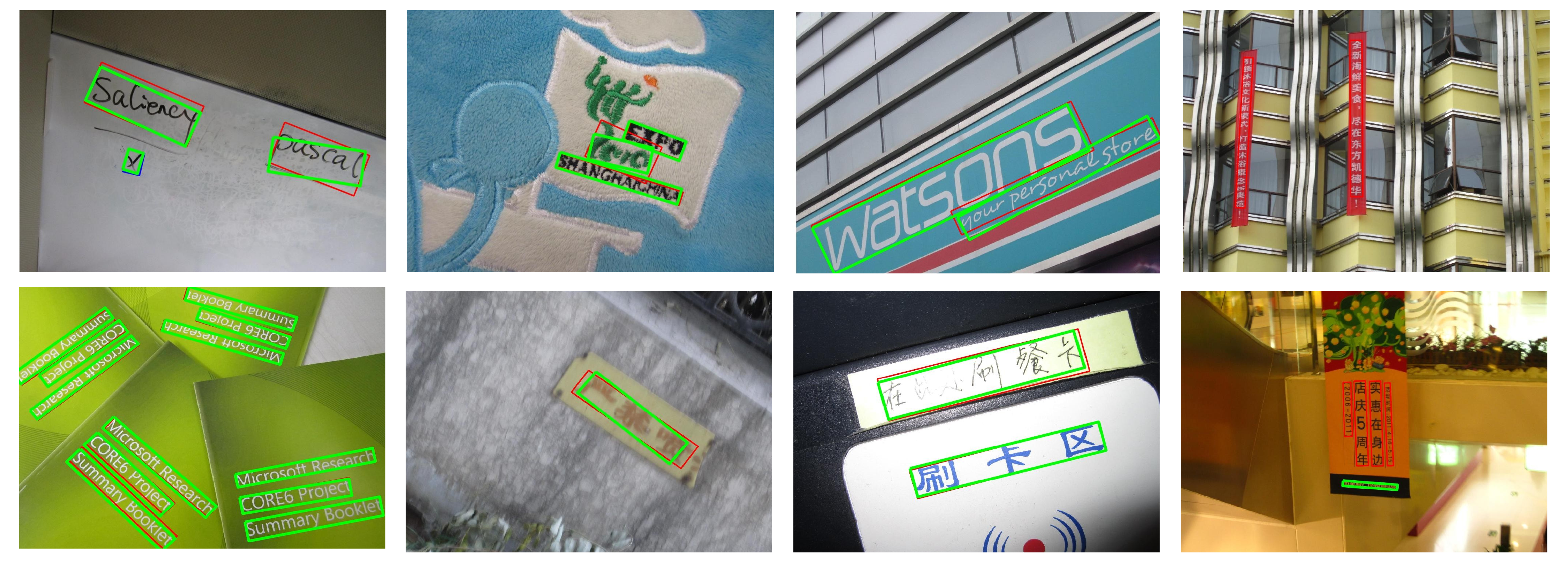}
\end{center}
\caption{For the sample images from the MSRA-TD500 dataset \cite{Re3}, the figure shows that the proposed technique can work for scene texts in arbitrary orientations and different languages in most cases (The red oriented rectangles are ground truth boxes and the green oriented rectangles are proposals.). It fails in rare cases when scene texts are vertically oriented as shown in the two images in the last column.}
\centering
\label{Orientfig_001}
\end{figure}

The proposed technique also can detect scene texts in different orientations and languages. We demonstrate this capability using the MSRA-TD500 dataset \cite{Re3} that contains scene texts in arbitrary orientations and a mixture of English and Chinese. Experiments show that recalls of 88.14\%, 83.33\% and 75.77\% are obtained under the IoU thresholds of 0.5, 0.7 and 0.8. These recalls are comparable to those achieved over the ICDAR2015 and SVT datasets (as shown in Tables \ref{table1} and \ref{table2}), where most texts are almost horizontal and printed in English. Fig. \ref{Orientfig_001} shows several sample images from the MSRA-TD500 that capture English and Chinese texts in arbitrary orientations, as well as text proposals by our proposed technique. As Fig. \ref{Orientfig_001} shows, the proposed technique is capable of detecting English and Chinese texts when there exist certain overlaps between neighbouring characters in the vertical direction. It fails when scene texts are vertically oriented as shown in the two images in the last column. Note that a maximum of 2000 proposals are generated in each image and the proposals shown in Fig. \ref{Orientfig_001} are those having the best overlapping with the ground truth.
\par

The superior performance of the MPT is largely attributed to the proposed pooling based grouping strategy that captures the exact text layout and appearance in scenes, i.e. characters are usually closer to each other (as compared with neighbouring non-text objects) forming words and text lines. In fact, the proposed grouping strategy can also handle texts with broken edges as far as they have certain overlap in the vertical direction. As a comparison, the EdgeBox (EB) \cite{EB} makes use of image edges similarly with a much lower recall rate, largely due to different grouping strategies. Besides the proposed grouping strategy, the HoGe based proposal ranking helps to shift scene text proposals to the front of the sorted list which also contributes to the superior performance of the proposed MPT technique when a limited number of proposals are selected.

\subsection{End-to-end and word spotting}
\label{E2EEval}

\begin{table}[!t]
\small
\caption{Word spotting performance of the developed end-to-end scene text reading system (MPT) and other state-of-the-art systems, including proposal based systems (*) and CNN-based systems for the ICDAR2015 dataset and the SVT dataset.}
\begin{center}
\begin{tabular}{c c cccccc}
\hline
\hline
\multirow{3}{*}{Method} & \multirow{3}{*}{SVT-50} & \multicolumn{6}{c}{Words Spotting}\\[-0.05cm]
\cline{3-8}
                         &                        & \multicolumn{3}{c}{strong} & \multicolumn{3}{c}{weak}\\[-0.05cm]
\cline{3-8}
& & R & P & F & R & P & F \\[-0.05cm]
\hline
\hline
EB\_Sys$^*$           & 72.84 & 55.26 & 66.81 & 60.49 & 54.79 & 57.48 & 56.10 \\[-0.05cm]
STL\_Sys$^*$          & 75.02 & 61.80 & 85.32 & 71.68 & 61.45 & 81.30 & 69.99 \\[-0.05cm]
TP\_Sys$^*$           & 76.41 & 66.47 & 89.47 & 79.27 & 65.07 & 82.40 & 72.72 \\[-0.05cm]
Jar-E2E\cite{JarE2E}  & 68    & 86.68 & 94.64 & 90.49 &   -   &   -   &   -   \\[-0.05cm]
ConvLSTM \cite{E2E11} & $-$   & 84.93 & \textbf{98.91} & 91.39 & 84 & \textbf{97.29} & 90.16\\[-0.05cm]
DeepTextSpotter\cite{DTSpotter} &$-$ & $-$ & $-$ & 92 & $-$ & $-$ & 89 \\[-0.05cm]
TextBoxes \cite{TextBoxes} & 84 & \textbf{90.77} & 97.25 & \textbf{93.90} & 87.38 & 97.02 & \textbf{91.95} \\[-0.05cm]
\hline
\textbf{Proposed MPT} & 84 & 88.20 & 97.55 & 92.64 & \textbf{87.85} & 95.31 & 91.43 \\[-0.05cm]
\hline
\hline
\end{tabular}
\end{center}
\label{T3}
\end{table}

\begin{table}[!t]
\small
\caption{End-to-End performance of the developed scene text reading system (MPT) and other state-of-the-art systems, including proposal based systems (*) and CNN-based systems for the ICDAR2015 dataset.}
\begin{center}
\begin{tabular}{c cccccc}
\hline
\hline
\multirow{3}{*}{Method} & \multicolumn{6}{c}{End-to-end Scene Text Reading}\\[-0.05cm]
\cline{2-7}
                        & \multicolumn{3}{c}{strong} & \multicolumn{3}{c}{weak} \\[-0.05cm]
\cline{2-7}
& R & P & F & R & P & F \\[-0.05cm]
\hline
\hline
EB\_Sys$^*$           & 52.67 & 65.80 & 58.51 & 52.24 & 56.69 & 54.37 \\[-0.05cm]
STL\_Sys$^*$          & 59.65 & 85.07 & 70.13 & 59.43 & 81.10 & 68.60 \\[-0.05cm]
TP\_Sys$^*$           & 63.90 & 88.12 & 74.08 & 62.70 & 81.21 & 70.77 \\[-0.05cm]
Jar-E2E \cite{JarE2E} & 82.12 & 91.05 & 86.35 &   -   &   -   &   -   \\[-0.05cm]
ConvLSTM \cite{E2E11} & 79.39 & \textbf{96.68} & 87.19 & 79.28 & 94.91 &86.39\\[-0.05cm]
DeepTextSpotter\cite{DTSpotter}& $-$ & $-$ & 89 & $-$ & $-$ & 86 \\[-0.05cm]
TextBoxes\cite{TextBoxes}      & \textbf{87.68} & 95.83 & \textbf{91.57} & \textbf{84.51} & \textbf{95.44} & \textbf{89.65} \\[-0.05cm]
\hline
\textbf{Proposed MPT} & 84.08 & 96.25 & 89.76 & 83.86 & 93.89 & 88.59 \\[-0.05cm]
\hline
\hline
\end{tabular}
\end{center}
\label{T4}
\end{table}

Tables \ref{T3} and \ref{T4} compare our developed end-to-end system with several state-of-the-art end-to-end scene text reading systems including several CNN-based: Jar-E2E model \cite{JarE2E}, ConvLSTM \cite{E2E11}, DeepTextSpotter\cite{DTSpotter}, and TextBoxes \cite{TextBoxes} as well as several proposal based: EB\_Sys, TP\_Sys, STL\_Sys which are constructed by combining EB, TP, and STL based scene text proposal techniques with Jarderberg's scene text recognition model. The comparisons are based on precision, recall and f-measure on the ICDAR2015 dataset and the SVT dataset. As the two tables show, the performance of the proposed system is clearly better than other proposal based systems and also comparable to the CNN-based systems. Note that the TextBoxes \cite{TextBoxes} trains two dedicated networks for detection and recognition, and it was trained using a huge amount images including images in the SynthText \cite{SynText} (containing 800,000 images) as well as training images in the ICDAR2011 dataset and the ICDAR2013 dataset. As a comparison, our proposed system was trained using 479 training images in the ICDAR2003 dataset and the ICDAR2013 dataset only.
\par 

\begin{figure}
\begin{center}
\includegraphics[width=4.8in]{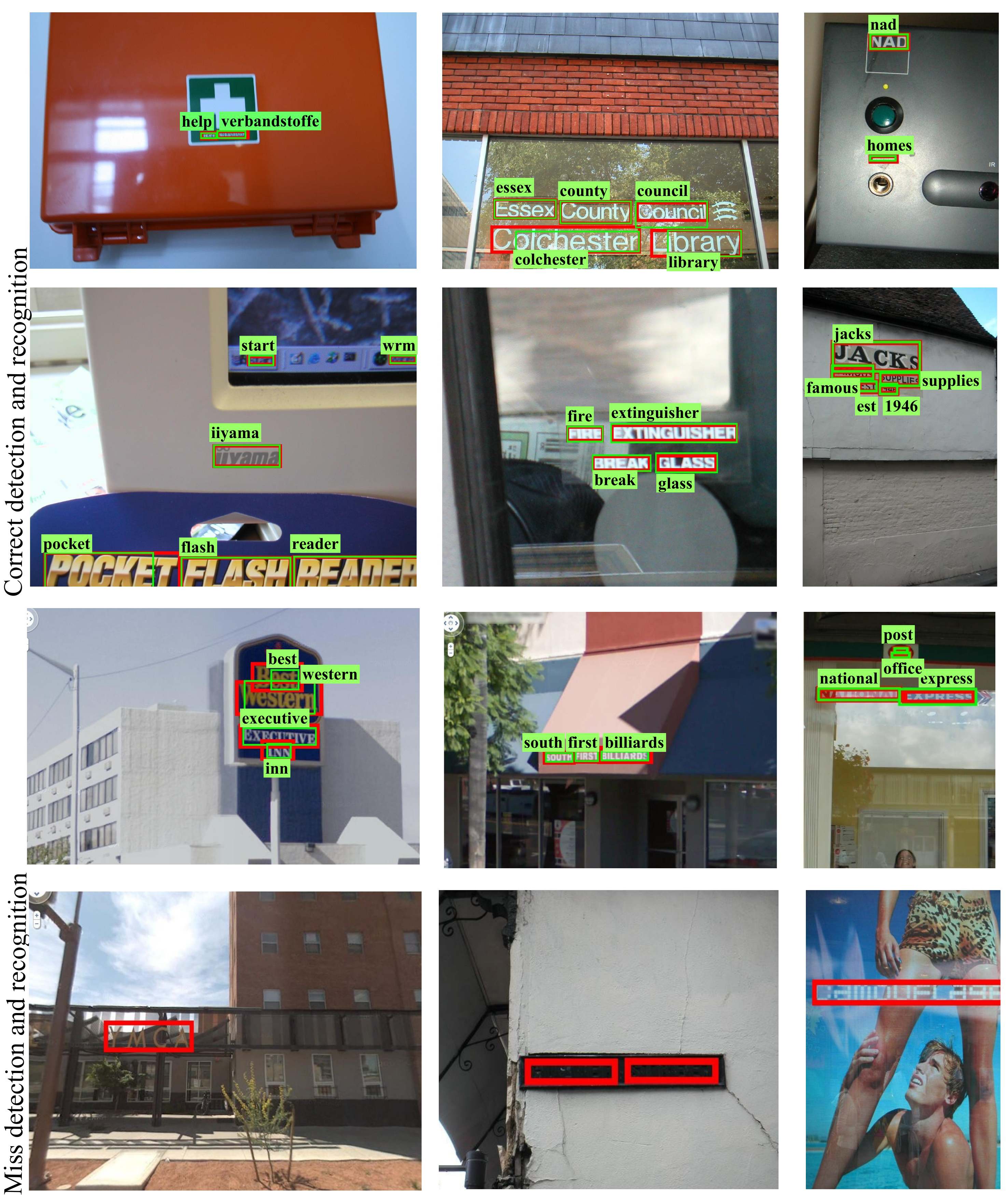}
\end{center}
\caption{
Several scene text detection and recognition examples where the proposed scene text reading system succeeds (the first three rows) and fails (the last row): The red boxes are ground truth and the green boxes are detection boxes by our proposed technique. The boxes with green-color background give the recognition results (words containing less than three characters are ignored).}
\centering
\label{fig_006}
\end{figure}

Fig. \ref{fig_006} shows a number of sample images that illustrate the performance of our developed end-to-end scene text reading system. As Fig. \ref{fig_006} shows, the proposed technique is capable of detecting and recognizing challenging texts with small text size (the first image in the first row), poor illumination and motion blur (the second images in the first and second rows), perspective distortion (the second image in the third row and the third image in second row). The superior scene text reading performance is largely due to the robustness of the proposed scene text proposal technique and the integrated scene text recognition model. Note that the proposed technique may fail when scene texts have ultra-low contrast or are printed in certain odd styles as illustrated in the sample images in the last row.

\section{Conclusion}
This paper presents a pooling based scene text proposal technique as well as its application to end-to-end scene text reading. The scene text proposal technique is inspired by the CNN pooling layer which is capable of grouping image edges into words and text lines accurately and efficiently. A novel score function is also designed which is capable of ranking generated proposals according to their probabilities of being text and accordingly helps to reduce the number of false-alarm proposals greatly. Further, the proposed  proposal technique does not rely on those heuristic thresholds/parameters such as text sizes, inter-character distances, etc. that are widely used in many existing techniques. Extensive experiments show that the pooling based proposal technique achieves superior performance as compared with state-of-the-arts. In addition, the integration of the pooling based proposal technique into an end-to-end scene text reading system also demonstrates state-of-the-art scene text reading performance.

\section*{Authors' Biographies}
\textbf{Dinh} is a PhD student at Sorbonne University – University Pierre and Marie CURIE, France. His current research works are in Image \& Pervasive Access Lab (IPAL, UMI2955, CNRS) and have collaboration with Institute for Infocomm Research (I2R, A-STAR), Singapore. His major research interests are visual understanding and machine learning.
\par
\textbf{Shijian} is an Assistant Professor with School of Computer Science \& Engineering, the Nanyang Technological University, Singapore. His major research interests include image and video analytic, visual intelligence, and machine learning. He published more than 80 international  journals and conference papers and co-authored over 10 patents in these research areas. 
\par
\textbf{Shangxuan-Tian} is a Senior Researcher at Tencent, China. Previously, he worked as a Research Scientist in the Institute for Infocomm Research, Singapore. He received his Ph.D. degree in School of Computing, National University of Singapore. His research interests include object detection and recognition, text understanding in scene images.
\par
\textbf{Nizar-Ouarti} is Associate Professor at Sorbonne UPMC in France. He is recently in CNRS delegation in the IPAL laboratory in Singapore. He received a PhD of College de France in 2007. He was postdoctoral researcher at INRIA. His topics of interest are ego-motion, computer vision and robotics.
\par
\textbf{Mounir-Mokhtari} is a Professor at Institut MINES TELECOM , France, Director of IPAL-CNRS French-Singaporean joint lab, Singapore, and Research Associate at CNRS-LIRMM Montpellier, France. His background is in human-machine interaction in the domain of Ambient Assistive Living. He has over 100 publications in journals, books and international conferences.

\section*{References}


\end{document}